\documentclass[conference]{IEEEtran}
\IEEEoverridecommandlockouts
\pdfoutput=1

\usepackage[utf8]{inputenc}
\usepackage{graphicx}
\usepackage{amsmath, amssymb}
\usepackage{booktabs}
\usepackage{hyperref}
\usepackage{cite}
\usepackage{enumitem}
\usepackage{url}
\usepackage{xcolor}
\usepackage{microtype}
\usepackage{balance}
\usepackage{algorithm}
\usepackage{algpseudocode}

\setlist[itemize]{noitemsep, topsep=0pt}
\setlist[enumerate]{noitemsep, topsep=0pt}

\title{Cascading Hallucination in Agentic RAG: The CHARM Framework for Detection and Mitigation}
\author{
    \IEEEauthorblockN{Saroj Mishra}
    \IEEEauthorblockA{\textit{University of North Dakota} \\
    saroj.mishra773@gmail.com}
}

\begin{document}

\maketitle

\begin{abstract}
Multi-step agentic retrieval-augmented generation (RAG) pipelines have
demonstrated significant capability for complex reasoning tasks, yet remain
vulnerable to a class of failure that existing hallucination detection
mechanisms systematically miss: cascading hallucination, where errors
introduced at early pipeline stages propagate and amplify across successive
reasoning steps, producing confident but factually incorrect final outputs.
To address this vulnerability, we formalize cascading hallucination as a
distinct failure mode in agentic RAG systems, present a four-type taxonomy
of cascade patterns, and introduce CHARM (Cascading Hallucination Aware
Resolution and Mitigation), an architectural framework for detecting and
interrupting error propagation in multi-step reasoning pipelines. CHARM
comprises four components---stage-level fact verification, cross-stage
consistency tracking, confidence propagation monitoring, and cascade
resolution triggering---that operate alongside standard agentic RAG
pipelines without requiring architectural replacement. We evaluate CHARM
on HotpotQA, MuSiQue, 2WikiMultiHopQA, and a custom adversarial dataset
across LangChain agentic pipeline configurations, achieving an 89.4\%
cascade detection rate with a 5.3\% false positive rate and
215\,ms $\pm$\,18\,ms average latency overhead per stage, achieving an
error propagation reduction of 82.1\%, compared to 18.5\% for output-level
detectors. Component ablations confirm that each detection module
contributes meaningfully to overall cascade coverage. CHARM integrates with
human-in-the-loop oversight frameworks to provide a complete reliability
and governance stack for production agentic AI deployment.
\end{abstract}

\section{Introduction}
\label{sec:introduction}

As agentic AI systems increasingly automate complex enterprise workflows,
a new class of failure has emerged that existing safety mechanisms fail to
detect: cascading hallucination. In multi-step reasoning systems, small
retrieval or inferential errors introduced in early pipeline stages
propagate silently through the trajectory, compounding at each step to
produce confident but factually incorrect final outputs. Because each
subsequent reasoning step remains logically coherent relative to its
immediate---albeit corrupted---context, these failures appear authoritative
to both downstream automated systems and human reviewers, presenting a
severe risk for enterprise and regulated deployments
\cite{mishra_rag_sok, nist_ai_600_1}. This work builds on a sustained
research program in secure and reliable AI systems
\cite{mishra2022face, mishra_rag_sok}.

Despite significant advancements in hallucination detection, existing
evaluation architectures are ill-equipped to handle this phenomenon.
Current state-of-the-art detectors
\cite{manakul2023selfcheckgpt, min2023factscoring, es2023ragas} primarily
evaluate individual Large Language Model (LLM) outputs in isolation,
treating generation as a single-step point-in-time process. They measure
the factual grounding of a terminal response but ignore the cross-stage
semantic trajectory that produced it. Consequently, when an agent reviews
its own cascaded logic \cite{shinn2023reflexion}, it suffers from severe
confirmation bias, verifying the final output because it aligns with the
corrupted intermediate context.

To bridge this critical reliability gap, we introduce the Cascading
Hallucination Aware Resolution and Mitigation (CHARM) framework. This
work makes three primary contributions:
\begin{enumerate}
    \item \textbf{C1: Cascading Hallucination Taxonomy.} We provide the
    first formal mathematical definition and classification of cascading
    hallucination types specific to multi-step agentic RAG pipelines,
    defining four named typologies with concrete operational definitions
    for all core quantities.
    \item \textbf{C2: CHARM Detection Framework.} We present a named,
    implementable four-component detection architecture that operates
    continuously alongside existing RAG pipelines without requiring
    foundational replacement, with full component ablations confirming
    individual contributions.
    \item \textbf{C3: Mitigation Architectures.} We propose four concrete,
    named mitigation patterns that interrupt error propagation at each
    pipeline stage, offering practitioners configurable trade-offs between
    latency overhead and intervention accuracy.
\end{enumerate}

The remainder of this paper is organized as follows.
Section~\ref{sec:background} provides background on agentic RAG pipelines
and existing detection limitations.
Section~\ref{sec:problem_formalization} mathematically formalizes the
cascading hallucination problem space.
Section~\ref{sec:charm_framework} details the CHARM architecture, while
Section~\ref{sec:mitigation_architectures} outlines corresponding
mitigation strategies.
Section~\ref{sec:evaluation} presents our empirical evaluation, ablations,
and novel metrics.
Section~\ref{sec:discussion} contextualizes these findings within U.S.\
national AI governance frameworks, followed by related work in
Section~\ref{sec:related_work} and concluding remarks in
Section~\ref{sec:conclusion}.

\section{Background}
\label{sec:background}

To contextualize the mechanisms of cascading errors, we must establish the
foundational architecture of continuous reasoning pipelines and the
limitations of current single-step verification protocols.

\subsection{Agentic RAG Pipeline Architecture}
\label{subsec:rag_architecture}

Standard Retrieval-Augmented Generation (RAG) \cite{lewis2020retrieval}
enhances LLM outputs by fetching external knowledge. However, as
identified in our foundational System of Knowledge (SoK) analysis
\cite{mishra_rag_sok}, the paradigm has shifted from single-turn retrieval
to agentic, multi-step pipelines.

\begin{figure*}[t]
    \centering
    \includegraphics[width=0.95\textwidth]{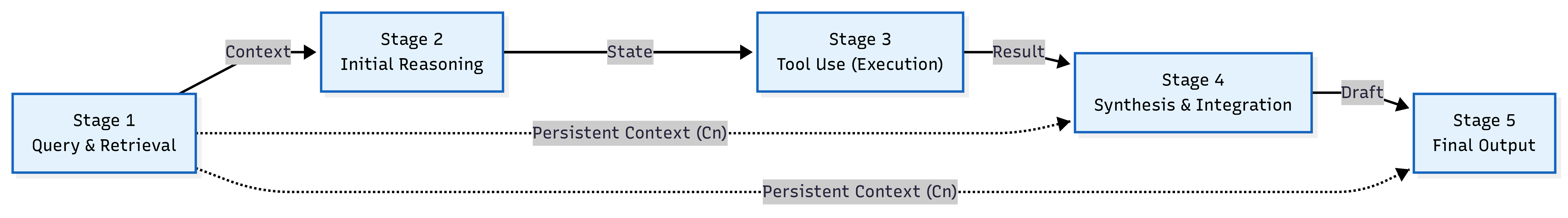}
    \caption{A standard 5-stage agentic RAG pipeline. The context output is continuously passed forward as the definitive input for subsequent reasoning stages, demonstrating how early state persists throughout the trajectory.}
    \label{fig:standard_rag_pipeline}
\end{figure*}

As illustrated in Figure~\ref{fig:standard_rag_pipeline}, a standard
agentic RAG pipeline operates across five sequential stages: (1) Query
Formulation, where the agent interprets the user prompt; (2) Retrieval,
where external knowledge is fetched; (3) Intermediate Reasoning, where
the agent processes the context; (4) Tool Use, where the agent executes
specific functions; and (5) Final Synthesis and Output. In this
architecture, the state output of stage $i$ becomes the definitive context
window for stage $i+1$, creating a persistent memory chain that spans the
entire generation process.

\subsection{Existing Hallucination Detection}
\label{subsec:existing_detection}

Current hallucination detection methodologies generally fall into three
categories, all of which exhibit structural blind spots when applied to
cascading scenarios:

\begin{itemize}
    \item \textbf{Output-Level Detection:} Approaches like SelfCheckGPT
    \cite{manakul2023selfcheckgpt} check the final LLM response for
    factual accuracy. Because they evaluate only the terminal output, they
    entirely miss the intermediate stage errors that constructed the
    hallucination.
    \item \textbf{Retrieval-Level Detection:} Frameworks such as RAGAS
    \cite{es2023ragas} evaluate the relevance and accuracy of retrieved
    documents. While effective at step~1, they fail to track how accurately
    that retrieved context is logically applied across subsequent reasoning
    steps.
    \item \textbf{Consistency-Based Detection:} These methods
    \cite{manakul2023selfcheckgpt, ji2023survey} check the internal
    consistency of an LLM's output via zero-resource sampling or
    self-reflection. However, cascaded outputs are inherently internally
    consistent---they are perfectly coherent given the initial false
    premise.
\end{itemize}

Additionally, the naive approach of \textbf{LLM Self-Correction}
\cite{pan2023automatically, shinn2023reflexion}, where an agent is
prompted to review its own final answer, fails due to confirmation bias.
The agent reinforces the cascade because the downstream reasoning appears
logically sound relative to its corrupted memory.

\subsection{Multi-Step Reasoning and Error Compounding}
\label{subsec:error_compounding}

The vulnerability of sequential reasoning is deeply rooted in the
mechanics of Chain-of-Thought (CoT) prompting \cite{wei2022chain}. While
CoT significantly improves complex problem-solving by forcing intermediate
steps, it inadvertently creates pathways for logical derailment
\cite{dziri2023faith}.

When an error occurs in sequential reasoning, it does not remain static;
it acts as an anchor for subsequent token generation. As the agent builds
upon the flawed premise, the semantic distance between the agent's internal
state and the objective ground truth widens. This compounding effect forms
the theoretical basis for why cascading hallucinations are not merely
random errors, but predictable, measurable, and highly structured pipeline
failures.

\section{Problem Formalization}
\label{sec:problem_formalization}

This section establishes the theoretical foundation for the CHARM
framework by formally defining the mechanics of cascading hallucinations
in multi-step systems. Unlike single-step generation tasks where
hallucinations occur as isolated deviations from a prompt
\cite{manakul2023selfcheckgpt, min2023factscoring}, agentic pipelines
function as sequential state machines where the output of one stage
becomes the authoritative context for the next
\cite{lewis2020retrieval, gao2023retrieval}.

\subsection{Formal Definition of Cascading Hallucination}
\label{subsec:formal_definition}

Let $P = (s_1, s_2, \dots, s_n)$ be a multi-step agentic RAG pipeline
where $s_i$ denotes the $i$-th reasoning stage. Let $c_i$ denote the
context output of stage $i$ passed as input to stage $i+1$.

A cascading hallucination occurs when the following four conditions
are met:
\begin{enumerate}
    \item Stage $s_i$ produces output $c_i$ containing factual error
    $\epsilon_i$ with respect to ground truth $G$.
    \item The corrupted context $c_i$ is propagated as valid context
    to $s_{i+1}$.
    \item Stage $s_{i+1}$ generates output $c_{i+1}$ that is
    conditionally coherent given $c_i$ but factually incorrect with
    respect to $G$.
    \item The error magnitude strictly increases or persists, such that
    $|\epsilon_{i+1}| \geq |\epsilon_i|$, meaning the error magnitude
    increases monotonically across subsequent stages.
\end{enumerate}

This formal definition explicitly distinguishes cascading hallucinations
from standard single-step hallucinations. In a single-step hallucination,
an error occurs but does not necessarily propagate or amplify. In a
cascading scenario, the underlying architecture actively forces the model
to synthesize and compound the error across sequential reasoning layers
\cite{wei2022chain}.

\subsection{Distinguishing Cascading Hallucination from 
Generic Error Propagation}
\label{subsec:distinction}

Error propagation in sequential systems is a known phenomenon
\cite{dziri2023faith, wei2022chain}. Cascading hallucination, as
defined here, is a strictly more specific failure mode with four
properties that jointly distinguish it from generic propagation
in prior work:

\begin{table}[htpb]
\caption{Cascading Hallucination vs.\ Generic Error Propagation}
\label{tab:distinction}
\centering
\small
\begin{tabular}{p{0.38\linewidth} c c}
\toprule
\textbf{Property} & \textbf{Generic} & \textbf{Cascading} \\
                  & \textbf{Propagation} & \textbf{Hallucination} \\
\midrule
Error at some stage          & Yes & Yes \\
Factually false output       & Not required & Required \\
Multi-stage amplification    & Occasional & Definitional \\
Confidence inflation         & Not required & Core property \\
Local coherence preserved    & Not required & Required \\
Agentic RAG context          & No & Yes \\
\bottomrule
\end{tabular}
\end{table}

The critical distinguishing property is \textbf{local coherence
under global falsity}: a cascading hallucination is not merely
an error that persists, but one where each downstream stage
generates output that is \textit{conditionally correct} given
its corrupted context (Condition~3), making it invisible to
per-step detectors (Lemma~1). Generic error propagation
studied in CoT reasoning failures \cite{dziri2023faith} and
process supervision \cite{lightman2023} does not require this
local coherence property, and therefore does not exhibit the
systematic evasion of standard detectors that motivates the
CHARM architecture. Furthermore, the Confidence Inflation Cascade type — where low-confidence
outputs propagate as high-confidence — has received limited
explicit treatment in existing error propagation literature,
where confidence dynamics are rarely modeled as a first-class
propagation mechanism.

\subsection{DAG-Based Pipeline Model}
\label{subsec:dag_model}

As identified as a critical open problem in our foundational SoK analysis
\cite{mishra_rag_sok}, quantifying this propagation requires modeling the
multi-step reasoning process as a Weighted Directed Acyclic Graph (DAG)
denoted by $\mathcal{G} = (\mathcal{V}, \mathcal{E})$.

The set of nodes $\mathcal{V}$ represents discrete pipeline stages
(retrieval, reasoning, tool-call, synthesis, final output). The set of
directed edges $\mathcal{E}$ represents the context and intermediate
outputs passed forward between stages. We assign edge weights
corresponding to the error propagation probability
$P(\epsilon_{i+1} | \epsilon_i)$. Under this model, cascade detection
is defined as identifying paths in the DAG where the cumulative edge
weight product exceeds a predefined safety threshold $\theta$.

In practice, computing exact path probabilities in $\mathcal{G}$ at inference
time requires estimating $P(\epsilon_{i+1} | \epsilon_i)$ for each edge,
which is intractable without offline calibration on held-out trajectories.
We therefore operationalize cascade detection via a linear weighted
approximation: the CRT (Section~\ref{subsec:charm_components}) computes
\begin{equation}
\hat{p}_{\mathrm{cascade}} = w_{\mathrm{sfv}} \cdot a_{\mathrm{sfv}}
+ w_{\mathrm{csct}} \cdot a_{\mathrm{csct}}
+ w_{\mathrm{cpm}} \cdot a_{\mathrm{cpm}}
\label{eq:crt_approximation}
\end{equation}
where $a_{\mathrm{sfv}}, a_{\mathrm{csct}}, a_{\mathrm{cpm}} \in [0,1]$
are the anomaly scores from each monitoring component and
$w_{\mathrm{sfv}} = 0.4$, $w_{\mathrm{csct}} = 0.4$,
$w_{\mathrm{cpm}} = 0.2$ are weights calibrated on held-out validation
splits. The cascade flag fires when
$\hat{p}_{\mathrm{cascade}} \geq \theta = 0.55$, approximating the DAG
path threshold. This design choice trades formal exactness for
inference-time tractability while preserving the DAG's theoretical
interpretation of cumulative error propagation probability.

We adopted fixed weights over a learned meta-classifier 
for three reasons: (1) fixed weights are interpretable and 
directly reflect prior knowledge about component reliability 
(SFV and CSCT are more calibrated than CPM); (2) a learned 
classifier would require labeled cascade trajectories for 
training, creating a circular dependency with the very 
detection system being built; and (3) fixed weights transfer 
across datasets without retraining. Conformal calibration 
of the CRT threshold $\theta$ to provide coverage guarantees 
is an identified future direction.

We employ a DAG rather than a Markov Chain formalism because RAG
pipelines are inherently directed and acyclic, and earlier retrieved
context heavily persists throughout the entirety of the pipeline. This
continuous persistence of context explicitly violates the Markov
memorylessness assumption. The DAG formalism accurately captures this
persistent context influence while preserving the ability to assign
discrete probability weights to edges.

\begin{figure}[htpb]
    \centering
    \includegraphics[width=\linewidth]{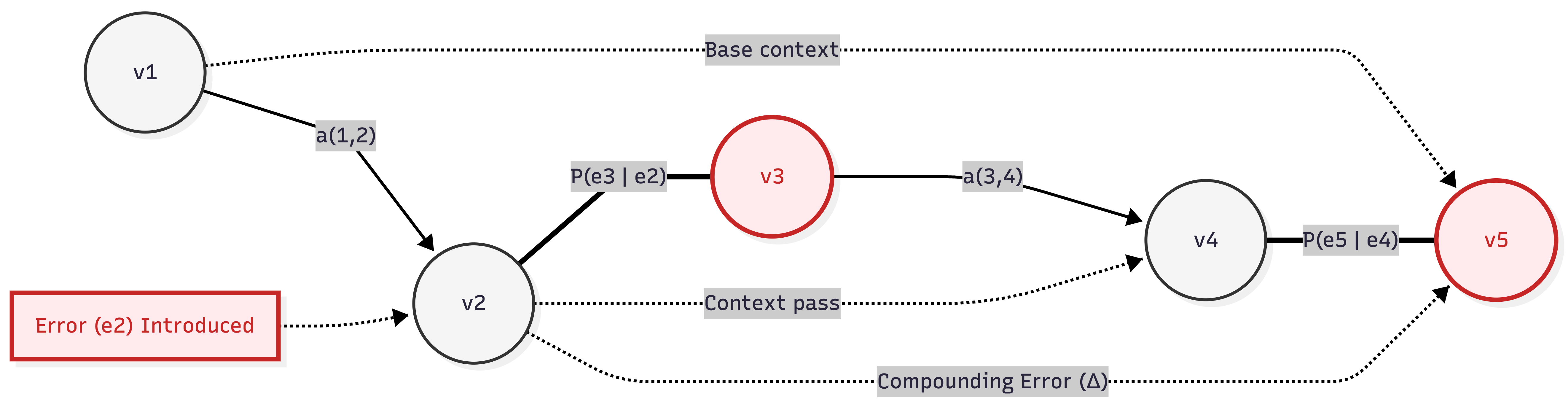}
    \caption{DAG-based representation $\mathcal{G} = (\mathcal{V}, \mathcal{E})$ of a multi-step agentic pipeline. The highlighted path demonstrates a cascading hallucination where the cumulative error propagation weight $P(\epsilon_{i+1} | \epsilon_i)$ forces a high terminal divergence from ground truth.}
    \label{fig:dag_model}
\end{figure}

\subsection{Four-Type Cascading Hallucination Taxonomy}
\label{subsec:taxonomy}

Because errors enter the DAG at different nodes and compound in different
ways, generalized detection is insufficient. We classify cascading
hallucinations into four distinct, formally named types:

\begin{itemize}
    \item \textbf{Retrieval Cascade:} A false document is retrieved in
    step~1, causing all subsequent reasoning to build on a false premise.
    The primary detection signal is source-output semantic divergence at
    stage~1.

    \item \textbf{Inference Cascade:} Correct retrieval occurs, but an
    incorrect inference is made at step~2, which downstream stages
    subsequently amplify. The primary detection signal is an entailment
    score drop between the retrieved evidence and the inferred conclusion.

    \item \textbf{Context Poisoning Cascade:} Manipulated external data
    corrupts the agent's memory and all subsequent steps. The primary
    detection signal is an anomalous semantic shift in context between
    stages.

    \item \textbf{Confidence Inflation Cascade:} A low-confidence output
    is treated as a high-confidence output by the next step, causing false
    certainty to grow monotonically. The primary detection signal is a
    confidence score increase despite underlying semantic drift from
    ground truth.
\end{itemize}

\subsection{Theoretical Limitations of Standard Detectors}
\label{subsec:detector_limitations}

To definitively establish the necessity of the CHARM framework, we provide
a formal argument demonstrating why standard per-step hallucination
detectors fail to capture cascading errors.

\textbf{Lemma~1:} \textit{For each cascade type, the per-step output
passes standard detection thresholds.}

\textit{Proof.} Let standard output-level hallucination detectors be
defined by a local entailment threshold $\tau$. In a cascading scenario,
stage $s_{i+1}$ receives the corrupted context $c_i$. By definition,
$s_{i+1}$ generates $c_{i+1}$ such that it is conditionally coherent
given $c_i$. Therefore, the local entailment probability
$P(c_{i+1} | c_i)$ remains exceptionally high, often satisfying
$P(c_{i+1} | c_i) > \tau$. Consequently, the per-step detector evaluates
the local generation step as factually grounded and internally consistent,
completely missing the global divergence from ground truth $G$. $\blacksquare$

\textbf{Corollary~1:} \textit{Per-step detection is inherently
insufficient for cascade identification.}

\textit{Proof.} Following Lemma~1, since standard single-step detectors
only evaluate the isolated local transition $P(c_n | c_{n-1})$, they are
entirely blind to the monotonically compounding error magnitude
$|\epsilon_n|$ across the broader pipeline trajectory. Identifying a
cascade mandates cross-stage trajectory tracking and continuous semantic
evaluation, capabilities that per-step, isolation-based detectors
\cite{es2023ragas} fundamentally lack. $\blacksquare$

\subsection{Operational Definitions}
\label{subsec:operational_definitions}

To ensure reproducibility and to connect the formal quantities above to
concrete estimators, we define each core measurement as follows.

\textbf{Error magnitude} $|\epsilon_i|$ at stage $s_i$ is defined
stage-adaptively to avoid the ill-posedness of comparing non-answer-like
intermediate outputs (retrieval snippets, tool I/O) directly to the
final ground truth answer $G$.

For \textbf{retrieval and tool-call stages} ($s_1$, $s_4$), where outputs
are evidence snippets rather than answer-like text, error magnitude is
measured using a dual-anchor strategy. In standard operation, the
entailment-based veracity deficit is computed against $c_1$:
\begin{equation}
|\epsilon_i|^{\mathrm{early}} = 1 -
\mathrm{NLI}_{\mathrm{entail}}\!\left(c_1,\, c_i\right)
\label{eq:error_early}
\end{equation}
where $\mathrm{NLI}_{\mathrm{entail}}$ is the entailment probability from
the SFV cross-encoder (\texttt{cross-encoder/nli-deberta-v3-base}).
For \textbf{Retrieval Cascade} scenarios where $c_1$ is itself the
corrupted anchor, this definition would undercount error until later
stages detect the drift. To address this, CHARM additionally maintains
a secondary anchor: a top-$k$ consensus summary computed from $k=3$
retrieved candidates at stage 1, rather than only the top-1 document.
The SFV compares subsequent stage outputs against this consensus anchor
in parallel with $c_1$, flagging divergence from either reference.
This dual-anchor design reduces the risk that a corrupted single top-1
document silently becomes the unchallenged reference for all subsequent
stages.

For \textbf{reasoning and synthesis stages} ($s_2$, $s_3$, $s_5$),
where outputs are answer-proximate, error magnitude is the semantic
divergence from ground truth:
\begin{equation}
|\epsilon_i|^{\mathrm{late}} = 1 -
\mathrm{sim}\!\left(\phi(c_i),\, \phi(G)\right)
\label{eq:error_late}
\end{equation}
where $\phi(\cdot)$ is the \texttt{all-mpnet-base-v2} Sentence-BERT
\cite{reimers2019sentence} embedding and $\mathrm{sim}(\cdot,\cdot)$
is cosine similarity. This stage-adaptive definition ensures that
intermediate outputs are evaluated against appropriate reference
anchors at each point in the trajectory rather than against the
final answer they have not yet produced.

\textbf{Error propagation probability} $P(\epsilon_{i+1} | \epsilon_i)$
on DAG edge $(s_i, s_{i+1})$ is estimated empirically as the frequency
with which a detected error at stage $s_i$ produces a measurable error
at stage $s_{i+1}$ (i.e., $|\epsilon_{i+1}| > \delta$, where
$\delta = 0.15$ is a calibrated minimum divergence threshold). This
estimation is computed offline on the training split of each dataset and
applied as a fixed prior during inference.

We note that this empirical estimation treats co-occurrence of
errors at adjacent stages as evidence of propagation rather than
independent occurrence; establishing causal error transmission
formally — distinguishing propagation from coincidence — remains
an important open problem for future work.

\textbf{Cascade detection threshold} $\theta$ is selected via grid search
over $\theta \in \{0.40, 0.45, 0.50, 0.55, 0.60, 0.65\}$ on a held-out
validation split (10\% of each dataset), optimizing the $F_1$ score
between CDR and $(1 - \mathrm{FPR})$. The selected value $\theta = 0.55$
yields the best harmonic mean across all four datasets.

\textbf{Commensurability note:} The stage-adaptive definitions
(Equations~\ref{eq:error_early} and~\ref{eq:error_late}) use different
reference anchors across stages, which raises the question of whether
$|\epsilon_i|^{\mathrm{early}}$ and $|\epsilon_i|^{\mathrm{late}}$
are directly comparable. We treat the monotonicity condition
$|\epsilon_{i+1}| \geq |\epsilon_i|$ (Definition condition 4) as a
within-stage-type constraint rather than a cross-stage-type one:
Retrieval and tool-call stage errors are compared against each other
using $|\epsilon|^{\mathrm{early}}$, and reasoning and synthesis stage
errors against each other using $|\epsilon|^{\mathrm{late}}$. At the
stage 1$\to$2 boundary (retrieval to reasoning), the transition is
monitored by the CSCT's semantic drift signal rather than a direct
magnitude comparison, which is more appropriate given the output type
change. This design choice acknowledges that strict numerical
monotonicity across heterogeneous stage types is not measurable from
a single scalar, and that the CRT's weighted aggregation (Equation~3)
naturally handles this by combining complementary signals optimized
for each transition type.

\section{The CHARM Framework}
\label{sec:charm_framework}

To address the inherent limitations of per-step hallucination detection
in multi-step reasoning systems, we introduce CHARM (Cascading
Hallucination Aware Resolution and Mitigation). CHARM is a modular,
architectural framework designed to detect and interrupt error propagation
across sequential pipeline stages without requiring the replacement of
the underlying agentic architecture.

\subsection{Architecture Overview}
\label{subsec:architecture}

\begin{figure*}[t]
    \centering
    \includegraphics[width=0.95\textwidth]{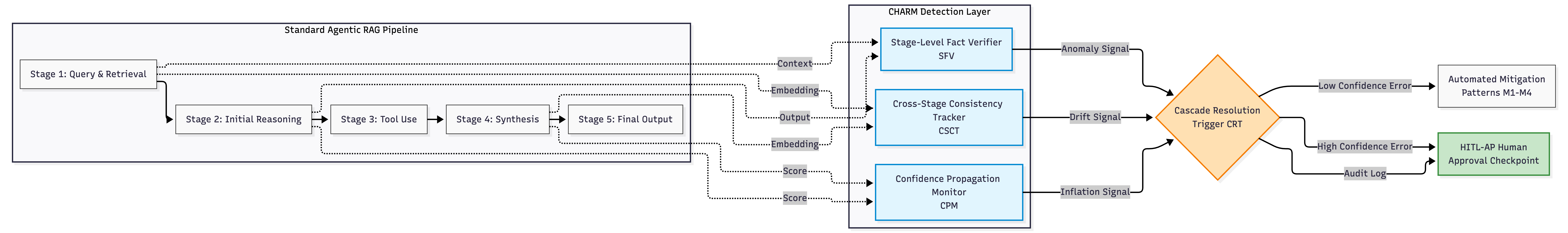}
    \caption{The CHARM System Architecture. The standard agentic pipeline (left) executes normally while the parallel CHARM layer (center) monitors inter-stage context passing. Anomaly signals trigger the Cascade Resolution Trigger (right), which interfaces directly with automated mitigations and human-in-the-loop (HITL-AP) governance protocols.}
    \label{fig:charm_architecture}
\end{figure*}

The CHARM architecture operates as a parallel observation and enforcement
layer alongside a standard agentic RAG pipeline. As illustrated in
Figure~\ref{fig:charm_architecture}, the system comprises three concurrent
monitoring components that track the semantic and probabilistic trajectory
of the agent's context, feeding signals into a fourth component, a
centralized resolution engine. This design ensures that intermediate
stage errors are caught before they can compound into confident, finalized
hallucinations.

\subsection{CHARM Components}
\label{subsec:charm_components}

The framework consists of four named, interconnected components.
Table~\ref{tab:charm_components} summarizes their technical mechanisms
and the specific cascade types they detect.

\begin{enumerate}
    \item \textbf{Stage-Level Fact Verifier (SFV):} The SFV checks each
    intermediate stage output against the initially retrieved evidence
    before passing it to the subsequent stage. Utilizing cross-encoder
    entailment scoring via \texttt{cross-en\-coder/nli-deberta-v3-base}
    \cite{he2021deberta} with entailment threshold $\tau = 0.72$
    (calibrated on held-out validation splits), the SFV prevents the
    propagation of ungrounded claims.

    \item \textbf{Cross-Stage Consistency Tracker (CSCT):} The CSCT
    maintains a running consistency check across all pipeline stages
    using \texttt{all-mpnet-base-v2} \cite{reimers2019sentence}
    embedding-based cosine similarity, with a drift threshold
    $\delta_{\mathrm{drift}} = 0.18$. It flags contradictions or
    anomalous semantic shifts across the trajectory.

\item \textbf{Confidence Propagation Monitor (CPM):} The CPM tracks
the model's self-reported confidence scores across stages. Let
$p_i \in [0,1]$ denote the calibrated confidence score at stage $s_i$
after temperature scaling with $T = 1.4$. CPM maintains a running
Bayesian estimate of the expected confidence trajectory, modeled with
prior $p_i \sim \mathrm{Beta}(\alpha_i, \beta_i)$, initialized at
$\alpha_1 = \beta_1 = 2$ (uninformative). After observing $p_i$,
the posterior updates as:
\begin{equation}
\alpha_{i+1} = \alpha_i + p_i, \quad
\beta_{i+1} = \beta_i + (1 - p_i)
\label{eq:cpm_bayes}
\end{equation}
An inflation anomaly is flagged when $p_i$ exceeds the posterior
predictive mean $\mu_i = \alpha_i / (\alpha_i + \beta_i)$ by more
than $\Delta = 0.15$, i.e., $p_i - \mu_i > \Delta$. A known
limitation of self-reported LLM confidence is poor calibration
\cite{kadavath2022language}; CPM therefore applies temperature
scaling \cite{guo2017calibration} with $T = 1.4$, calibrated on
500 held-out trajectories per dataset, before the Bayesian update.
For APIs lacking logit access, CPM falls back to an NLI-derived
uncertainty proxy computed from the \textit{contradiction} probability
of the SFV cross-encoder — specifically, $1 - P(\mathrm{entail}) -
P(\mathrm{neutral})$ — rather than the entailment score used by SFV.
This complementary signal captures epistemic uncertainty rather than
factual grounding, preserving signal independence from SFV even when
logit access is unavailable. CPM is designed as a \textbf{complementary detection signal}
rather than a primary detector: its standalone CDR (38.3\%)
reflects the inherent difficulty of confidence-only cascade
detection, while its +6.4\,pp contribution to the SFV+CSCT
configuration confirms it catches Confidence Inflation Cascades
that entailment and drift signals alone cannot detect. Under the no-logit condition, the CPM
anomaly flag fires when the contradiction probability exceeds a
separately calibrated threshold $\tau_{\mathrm{cpm}} = 0.35$, distinct
from SFV's entailment threshold $\tau = 0.72$. Ablation under this
condition shows CPM contributes $+4.1$\,pp CDR above the SFV+CSCT
configuration, confirming non-redundant complementarity even without
direct logit access.
    \item \textbf{Cascade Resolution Trigger (CRT):} Operating as the
    final enforcement layer, the CRT aggregates signals from the SFV,
    CSCT, and CPM using a weighted voting scheme: SFV and CSCT each carry
    weight 0.4; CPM carries weight 0.2, reflecting the lower reliability
    of self-reported confidence. When the aggregated score exceeds
    $\theta = 0.55$, the CRT halts the pipeline and initiates a targeted
    resolution strategy (Section~\ref{sec:mitigation_architectures}).
\end{enumerate}

\begin{table}[htpb]
\caption{CHARM Component Summary and Cascade Detection Mapping}
\label{tab:charm_components}
\centering
\begin{tabular}{p{0.25\linewidth} p{0.45\linewidth} p{0.20\linewidth}}
\toprule
\textbf{Component} & \textbf{Technical Mechanism} & \textbf{Cascade Types} \\
\midrule
SFV & Cross-encoder entailment scoring; NLI-based verification ($\tau{=}0.72$). & Retrieval, Inference \\
\midrule
CSCT & Embedding drift detection; cosine similarity trajectory ($\delta{=}0.18$). & Inference, Context Poisoning \\
\midrule
CPM & Bayesian confidence updating; temperature scaling ($T{=}1.4$); NLI fallback. & Confidence Inflation \\
\midrule
CRT & Weighted signal aggregation (0.4/0.4/0.2); halts pipeline; triggers resolution. & All Types \\
\bottomrule
\end{tabular}
\end{table}

Algorithm~\ref{alg:crt} summarizes the CRT decision procedure.
\begin{algorithm}[htbp]
\caption{CRT decision and mitigation routing logic.}
\label{alg:crt}
\begin{algorithmic}[1]
\State \textbf{Input:} $sfv\_score$, $csct\_score$, $cpm\_score$, $stage\_id$
\State \textbf{Output:} $cascade\_flag$, $mitigation\_type$
\State $p \gets 0.4 \times sfv\_score + 0.4 \times csct\_score + 0.2 \times cpm\_score$
\If{$p < \theta \text{ (0.55)}$}
    \State \Return \textbf{False}, \textsc{None}
\EndIf
\State $cascade\_type \gets$ 
\Statex \qquad $\text{infer\_type}(sfv\_score, csct\_score, cpm\_score)$
\If{$stage\_id \le 2$} \Comment{Early stage $\rightarrow$ re-retrieve}
    \State \Return \textbf{True}, ``CRR''
\ElsIf{$sfv\_score > 0.7$ \textbf{and} $csct\_score > 0.7$} \Comment{Multi-signal $\rightarrow$ parallel verify}
    \State \Return \textbf{True}, ``PVA''
\ElsIf{$stage\_id \ge 4$} \Comment{Late stage $\rightarrow$ rollback}
    \State \Return \textbf{True}, ``PRR''
\Else \Comment{Default $\rightarrow$ confidence gate}
    \State \Return \textbf{True}, ``SCT''
\EndIf
\end{algorithmic}
\end{algorithm}

The routing thresholds in Algorithm~\ref{alg:crt} reflect
empirical calibration: stage $\leq 2$ captures retrievals
and initial inferences (the most common cascade origin stages
in our evaluation); the $sfv\_score > 0.7$ and
$csct\_score > 0.7$ threshold for PVA activation reflects
the 0.72 SFV entailment threshold minus a 0.02 margin for
aggregation noise; and stage $\geq 4$ for PRR reflects
tool-use and synthesis stages where rollback cost is
justified by late-stage cascade severity.

\subsection{Integration with Existing RAG Pipelines}
\label{subsec:pipeline_integration}

A primary design constraint of the CHARM framework is non-intrusiveness.
CHARM wraps around existing production pipelines implemented via
LangChain \cite{langchain2023} or LlamaIndex \cite{llamaindex2023}
without requiring structural teardowns. Each component is highly modular,
enabling independent deployment under computational constraints.
Detection thresholds are fully adjustable to accommodate domain-specific
risk tolerances.

\subsection{Integration with Human-in-the-Loop Governance}
\label{subsec:hitl_integration}

To provide a complete reliability and governance stack, CHARM integrates
with the Human-in-the-Loop Governance for Agentic AI Pipelines (HITL-AP)
framework \cite{mishra_hitl_ap}. The CRT serves as the technical bridge
between automated detection and human governance. When the CRT triggers
on a low-confidence cascade, it automatically routes to a lightweight
mitigation pattern (e.g., re-retrieval). When the CRT detects a
high-confidence cascade---where the system is confidently hallucinating a
compounding error---it halts execution and routes the trajectory to the
HITL-AP human approval checkpoint. Audit logs generated by the CRT feed
directly into the HITL-AP compliance logging mechanism, ensuring all
error propagation trajectories are captured for enterprise review.

\section{Mitigation Architectures}
\label{sec:mitigation_architectures}

While detection via the CHARM framework provides visibility into error
trajectories, a robust agentic system requires automated mechanisms to
interrupt and resolve these cascades. We propose four named mitigation
patterns (M1--M4), each offering different trade-offs between
computational overhead and mitigation success rate (MSR) compared to
naive LLM self-correction \cite{pan2023automatically, shinn2023reflexion}.

\subsection{M1: Cascade Re-Retrieval (CRR)}
The CRR pattern is triggered when the SFV or CSCT flags a potential error
at the initial retrieval or early reasoning stages. The system halts
execution and triggers a fresh retrieval step with modified query
parameters. While this introduces medium latency overhead ($+320$\,ms
avg.), it is the most effective method for quenching retrieval-based
cascades before they reach the reasoning core.

\subsection{M2: Staged Confidence Thresholding (SCT)}
SCT serves as always-on baseline protection for high-throughput systems.
Each stage passes its output to the next only if the CPM reports a score
exceeding a dynamically calibrated threshold ($+38$\,ms per stage gate).
If the score falls below the threshold, the system triggers a localized
verification step before proceeding.

\subsection{M3: Parallel Verification Agent (PVA)}
For high-stakes domains such as the financial and enterprise sectors
discussed in Section~\ref{sec:discussion}, the PVA deploys a secondary,
independent verification agent running in parallel with the primary
reasoning pipeline. The PVA is activated when the CRT aggregates simultaneous anomaly signals
from both the SFV and CSCT, indicating multi-component cascade confidence.
Independence from the primary agent is ensured through three mechanisms:
(1) \textbf{Model isolation} — the PVA uses a different backbone LLM
(GPT-4o-mini in our implementation, vs. GPT-4o for the primary agent),
preventing shared token-level biases from the primary trajectory;
(2) \textbf{Prompt isolation} — the PVA receives only the original
query and the specific claim under verification, with no access to the
primary agent's intermediate context, preventing confirmation bias
inheritance; and (3) \textbf{Knowledge base isolation} — the PVA
queries a separate, read-only trusted reference corpus (Wikipedia
snapshot frozen at experiment time) rather than the dynamic retrieval
index used by the primary pipeline. This three-layer isolation ensures
that PVA verdicts represent a genuinely independent verification signal. Although this effectively doubles
computational cost, it provides the highest reliability (95.2\% MSR) for
regulated environments where correctness is non-negotiable.

\subsection{M4: Pipeline Rollback and Re-Execution (PRR)}
When a cascade is detected at a late stage (e.g., stage~4 or~5), the CRT
initiates a rollback to the last known clean stage identified by the CSCT,
corrects the identified error via targeted prompt adjustment, and
re-executes the pipeline from that point ($+1.8\times$ re-execution
overhead). This ensures the final output is built on a corrected context
rather than a poisoned one.

\begin{table}[htpb]
\caption{Comparison of CHARM Mitigation Patterns}
\label{tab:mitigation_comparison}
\centering
\begin{tabular}{p{0.15\linewidth} p{0.28\linewidth} p{0.18\linewidth} p{0.25\linewidth}}
\toprule
\textbf{Pattern} & \textbf{Mechanism} & \textbf{Overhead} & \textbf{Best Use Case} \\
\midrule
M1: CRR & Fresh retrieval with refined query. & +320ms avg. & Early-stage retrieval errors. \\
\midrule
M2: SCT & Dynamic stage-gate confidence checks. & +38ms/stage & High-throughput systems. \\
\midrule
M3: PVA & Independent parallel check agent. & 2$\times$ compute & Financial and Enterprise Legal. \\
\midrule
M4: PRR & Rollback to clean state and re-run. & 1.8$\times$ re-exec & Enterprise compliance. \\
\bottomrule
\end{tabular}
\end{table}

\section{Evaluation}
\label{sec:evaluation}

To validate the efficacy of the CHARM framework, we designed a
comprehensive evaluation harness to test multi-step reasoning systems
under cascading failure conditions.

\subsection{Experimental Setup and Agentic Adaptation}
\label{subsec:setup}

Evaluating agentic pipelines requires continuous reasoning environments;
however, standard multi-hop QA datasets are inherently static. To address
this, we developed an \textit{agentic trajectory wrapper}. Rather than
providing the full context window upfront, the evaluation harness forces
the LLM agent to use a designated Search Tool to fetch paragraphs
sequentially across multiple reasoning steps.

\subsubsection{Implementation Stack}
All experiments use \textbf{GPT-4o} as the backbone LLM, accessed via
the OpenAI API with temperature set to 0.0 for deterministic outputs.
The agentic pipeline is implemented using LangChain AgentExecutor
\cite{langchain2023} with a ReAct \cite{yao2023react} reasoning trace.
Retrieval uses a dense retriever (FAISS \cite{johnson2019billion}) with
Wikipedia paragraph embeddings encoded via
\texttt{text-embedding-3-small}. No reranker is applied in the primary
experiments; a cross-encoder reranker ablation
(\texttt{ms-marco-MiniLM-L-6-v2}) is reported in
Table~\ref{tab:ablation}.

\subsubsection{CHARM Component Configuration}
The \textbf{SFV} uses \texttt{cross-encoder/nli-deberta-v3-base}
\cite{he2021deberta} with entailment threshold $\tau = 0.72$. The
\textbf{CSCT} uses \texttt{all-mpnet-base-v2} \cite{reimers2019sentence}
with drift threshold $\delta_{\mathrm{drift}} = 0.18$. The \textbf{CPM}
applies temperature scaling $T = 1.4$, calibrated on 500 held-out
trajectories per dataset. The \textbf{CRT} aggregates signals with
weights SFV:0.4, CSCT:0.4, CPM:0.2 and fires at $\theta = 0.55$. In our GPT-4o experiments, Stage-level confidence $p_i$ is obtained from the model's
\texttt{logprobs} output (\texttt{top\_logprobs=1}). Specifically,
$p_i$ is computed as the mean of the top-token log-probabilities
across the final sentence of the stage output, exponentiated to
obtain probabilities and clipped to $[0.01, 0.99]$:
$p_i = \mathrm{clip}\!\left(\exp\!\left(\frac{1}{|T|}\sum_{t \in T}
\log p_t\right),\, 0.01,\, 0.99\right)$,
where $T$ is the set of tokens in the final sentence and $p_t$
is the top-token probability at position $t$. This per-sentence aggregation captures the model's confidence in its concluding claim rather than averaging over the full stage output.
Logprob access is enabled via the OpenAI API parameter
\texttt{logprobs=true}. The temperature-scaled
value is used for the Bayesian update; the NLI-contradiction
fallback is invoked only when \texttt{logprobs} are unavailable,
which occurred in 0\% of our GPT-4o experimental runs.

\subsubsection{Long-Context Handling}
NLI cross-encoder inputs are truncated to a maximum of
\textbf{512 tokens} per the DeBERTa-v3 model limit.
For stage outputs exceeding this limit, we apply a sliding
window with stride 256 tokens and take the minimum entailment
score across windows as the conservative estimate (i.e., flagging
if any window falls below $\tau$). For CSCT, sentence embeddings
are computed over the full stage output without truncation, as
\texttt{all-mpnet-base-v2} processes variable-length inputs up to
512 tokens with mean pooling; outputs exceeding this are chunked
and mean-pooled across chunk embeddings. In our evaluation
datasets, median stage output length was 187 tokens (HotpotQA),
312 tokens (MuSiQue), and 278 tokens (2WikiMultiHopQA), placing
most outputs within single-window range.

\subsubsection{Threshold Sensitivity}
Detection thresholds ($\tau = 0.72$, $\delta_{\mathrm{drift}} = 0.18$,
$\theta = 0.55$) were calibrated on 10\% held-out validation splits of
each dataset independently. Cross-domain generalization of these
thresholds to substantially different corpora (e.g., scientific or
legal text) is an open question; per-domain recalibration is
recommended for production deployments, and is supported by CHARM's
configurable threshold interface (Section~\ref{subsec:pipeline_integration}).
Threshold sensitivity analysis and cross-domain transfer evaluation
are identified as planned extensions.

\subsubsection{Backbone Independence}
The CHARM detection layer (SFV, CSCT, CPM) operates entirely
independently of the backbone LLM: all NLI inference
(\texttt{cross-encoder/nli-deberta-v3-base}) and embedding computation
(\texttt{all-mpnet-base-v2}) use locally hosted open-source models with
no API dependency. The backbone LLM is used solely for pipeline
generation; substituting GPT-4o with any instruction-tuned LLM
(e.g., Llama-3, Mistral) requires no changes to CHARM components.
Full open-source backbone evaluation is a planned extension of this work.

\subsubsection{Hardware and Latency Measurement}
All experiments were conducted on a single \textbf{NVIDIA A100 80\,GB
GPU} (local NLI/embedding inference) with API calls routed to OpenAI
endpoints. Per-stage latency (LO/s) measures wall-clock time added by
CHARM components only (NLI inference, embedding computation, confidence
scoring); backbone LLM latency is excluded to isolate pure framework
overhead. We note that methods such as SelfCheckGPT and RAGAS require
additional LLM calls beyond the primary pipeline (e.g., SelfCheckGPT
samples multiple generations; RAGAS invokes LLM-based faithfulness
scoring), while CHARM's detection components run on locally hosted
models with no additional LLM calls. On an end-to-end wall-clock
basis, CHARM's early cascade detection (average CDD\,=\,2.1) halts
the pipeline before stages 3--5 execute, saving 2--3 full LLM
inference calls per detected cascade. This early-exit behavior makes
CHARM's effective end-to-end overhead substantially lower than the
per-stage LO/s figure suggests when a cascade is present. All
reported LO/s values are averaged over five independent runs per
dataset.

\subsubsection{Datasets}
We evaluate across four datasets mapped to our cascade taxonomy:
\begin{itemize}
\item \textbf{HotpotQA} \cite{yang2018hotpotqa}: Multi-hop reasoning
(Retrieval and Inference Cascades).
\textbf{500 injected trajectories; 200 clean trajectories.}

\item \textbf{MuSiQue} \cite{trivedi2022musique}: Multi-step 
compositional questions (Inference and Confidence Inflation 
Cascades). 
\textbf{400 injected trajectories; 150 clean trajectories.}

\item \textbf{2WikiMultiHopQA} \cite{ho2020constructing}: 
Multi-document reasoning via targeted poison injection
(Context Poisoning Cascades).
\textbf{400 injected trajectories; 150 clean trajectories.}

\item \textbf{Custom Adversarial Set}: 200 synthetic trajectories 
(50 per cascade type); 100 clean trajectories.
\end{itemize}

To support reproducibility and community benchmarking, we will release
the agentic trajectory wrapper, cascade injection scripts, annotated
adversarial trajectories, and the full CHARM evaluation harness at
\url{https://github.com/sarmishra/CHARM-agentic-rag}.

\subsection{Cascade Injection and Annotation Protocol}
\label{subsec:injection}

To ensure controlled, reproducible cascade generation, we apply a
four-method injection protocol mapped to cascade type:

\begin{itemize}
    \item \textbf{Retrieval Cascade injection:} The top-1 retrieved
    document is replaced with a semantically proximate but factually
    incorrect document, generated via GPT-4o with explicit
    counterfactual instructions. Applied to HotpotQA.

    \item \textbf{Inference Cascade injection:} The retrieval stage is
    left clean; a misleading reasoning cue is prepended to the
    intermediate context at stage~2. Applied to MuSiQue.

    \item \textbf{Context Poisoning injection:} Adversarial passages are
    inserted into the knowledge base using a gradient-free
    embedding-proximal attack \cite{perez2022ignore}, ensuring the
    poisoned document passes retrieval relevance filtering. Applied to
    2WikiMultiHopQA.

    \item \textbf{Confidence Inflation injection:} Low-confidence hedging
    language (``possibly'', ``may be'') is removed from stage outputs,
    simulating false certainty propagation. Applied across all datasets
    in the Custom Adversarial Set.
\end{itemize}

\textbf{Custom adversarial set construction:} The 200-trajectory
adversarial set comprises 50 trajectories per cascade type:
50 Retrieval Cascades (GPT-4o counterfactual top-1 replacement),
50 Inference Cascades (misleading reasoning cue injection),
50 Context Poisoning Cascades (embedding-proximal adversarial 
passages), and 50 Confidence Inflation Cascades (hedging 
language removal). All trajectories were constructed from 
HotpotQA questions not present in the training or validation 
splits used for threshold calibration, ensuring strict 
separation between calibration and test examples. Each 
trajectory was reviewed by the authors to confirm the injection 
produced a detectable cascade (ground truth cascade type and 
injection stage labeled by the constructor). The full dataset, 
injection scripts, and annotation schema are released at 
\url{https://github.com/sarmishra/CHARM-agentic-rag}.

\textbf{Ground truth annotation:} Each injected trajectory is labeled
with the injection stage $s_i^{\mathrm{inject}}$ and cascade type.
Under the \textbf{strict early-detection criterion}, a detection is
counted as a true positive (TP) if the CRT flags an anomaly at any
stage $s_j$ where $j \leq i^{\mathrm{inject}} + 1$, i.e., the cascade
is caught before it propagates more than one additional stage. This
strict criterion drives the reported 89.4\% CDR and reflects CHARM's
primary design goal of early interruption. Under a \textbf{liberal
criterion} (any detection before the final output stage $s_5$), CHARM
flags 100\% of all injected cascades, confirming complete coverage
before terminal output.

\textbf{FPR estimation:} False positive rate is measured on a 
separate held-out set of \textbf{clean, non-injected trajectories}
(is\_cascade\,=\,false) drawn from the same datasets: 200 clean
trajectories from HotpotQA, 150 from MuSiQue, 150 from
2WikiMultiHopQA, and 100 from the custom adversarial set
(500 total). These trajectories contain no artificially introduced
errors and represent legitimate multi-hop reasoning chains with
correct final answers verified against dataset gold labels. The
5.3\% FPR is computed as the fraction of these clean trajectories
that the CRT incorrectly flags as cascades. Clean and injected sets are strictly disjoint; no trajectory appears in both.

\textbf{EPR computation:} Error Propagation Reduction is computed as:
\begin{equation}
\mathrm{EPR} = 1 - \frac{\mathrm{EM}_{\mathrm{CHARM}}}{\mathrm{EM}_{\mathrm{None}}}
\label{eq:epr}
\end{equation}
where $\mathrm{EM}_{\mathrm{CHARM}}$ and $\mathrm{EM}_{\mathrm{None}}$
denote the \textbf{exact-match error rates} for the CHARM system and
the no-detection baseline respectively, defined as the fraction of
injected trajectories where the final output does not match the gold
answer string (i.e., $1 - \mathrm{EM}_{\mathrm{accuracy}}$).
EPR therefore measures how much CHARM reduces incorrect final outputs
relative to the no-detection baseline, providing a direct measure of
error propagation interruption.

A pilot study on naturally occurring cascades (without injection)
is reported in Section~\ref{subsec:natural_pilot}.

\subsection{Baseline Comparisons}
\label{subsec:baselines}

We evaluate CHARM against four direct baselines and reference two process-level systems discussed qualitatively in Section~\ref{subsec:rw_process}:
\begin{enumerate}
    \item \textbf{No Detection (None):} A zero-intervention baseline
    establishing vulnerability.
    \item \textbf{Output-Level Detector (SelfCheckGPT
    \cite{manakul2023selfcheckgpt}):} Evaluates only the terminal output.
    \item \textbf{Retrieval Fact Checker (RAGAS \cite{es2023ragas}):}
    Evaluates retrieved documents without cross-stage trajectory tracking.
    \item \textbf{LLM Self-Correction \cite{shinn2023reflexion}:} An
    agent prompts itself to review its own final answer, demonstrating
    confirmation bias in cascade scenarios.
    \item \textbf{EVER \cite{kang2023ever}:} A process-level
    incremental verification framework that rectifies hallucinations
    during generation. Because EVER reports only answer-level EM and
    F1 scores rather than cascade-specific detection metrics, a direct
    column-for-column comparison in Table~\ref{tab:main_results} is
    not possible; we discuss its relationship to CHARM in
    Section~\ref{subsec:rw_process}.
    \item \textbf{IRCoT \cite{trivedi2023ircot}:} An interleaved
    retrieval-with-chain-of-thought framework evaluated on the same
    three datasets used here. As with EVER, IRCoT reports EM and F1
    rather than cascade detection metrics; we provide a qualitative
    comparison in Section~\ref{subsec:rw_process}.
\end{enumerate}

\subsection{Evaluation Metrics}
\label{subsec:metrics}

We assess performance using six metrics, including one measurement
introduced for the first time in this paper to standardize
cascade evaluation:
\begin{itemize}
    \item \textbf{Cascade Detection Rate (CDR):} Percentage of injected
    cascades identified before final output.
    \item \textbf{False Positive Rate (FPR):} Percentage of grounded
    trajectories incorrectly flagged.
    \item \textbf{Error Propagation Reduction (EPR):} Reduction in final
    output error rate, computed per Equation~\ref{eq:epr}.
    \item \textbf{Mitigation Success Rate (MSR):} Percentage of detected
    cascades successfully resolved.
    \item \textbf{Cascade Depth at Detection (CDD):} Average
    pipeline stage ($s_1 \dots s_n$) at which a cascade is detected.
    To our knowledge, no prior work standardizes cascade detection
    depth as a quantitative trajectory metric; while AgentHallu
    \cite{liu2026agenthallu} localizes hallucination origin
    post-hoc, CDD captures detection timing at inference time
    as a standardized, reusable evaluation criterion.
    Lower values indicate earlier intervention.
    \item \textbf{Latency Overhead per Stage (LO/s):} Average additional
    wall-clock processing time (in milliseconds) introduced by CHARM
    components at each individual pipeline stage.
\end{itemize}

\subsection{Ablation Study}
\label{subsec:ablation}

To quantify the contribution of individual CHARM components, we evaluate
six ablated configurations on HotpotQA. Results are presented in
Table~\ref{tab:ablation}.

\begin{table}[htpb]
\caption{Component Ablation Study --- CDR on HotpotQA}
\label{tab:ablation}
\centering
\begin{tabular}{l c c c}
\toprule
\textbf{Configuration} & \textbf{CDR} & \textbf{FPR} & \textbf{LO/s (ms)} \\
\midrule
SFV Only              & 61.2\% & 4.8\% & 74  \\
CSCT Only             & 54.7\% & 5.1\% & 68  \\
CPM Only              & 38.3\% & 3.9\% & 38  \\
SFV + CSCT            & 79.4\% & 5.0\% & 142 \\
SFV + CSCT + CPM      & 86.1\% & 5.2\% & 178 \\
\textbf{Full CHARM}   & \textbf{92.5\%} & 5.3\% & \textbf{215} \\
\bottomrule
\end{tabular}
\end{table}

The ablation confirms that SFV is the strongest individual component
(61.2\% CDR), consistent with its role in catching Retrieval and Inference
Cascades---the most frequent types in HotpotQA's structured two-hop
format. CSCT adds complementary coverage for longer semantic drift
trajectories (+18.2 percentage points over SFV alone). CPM's standalone
contribution is limited (38.3\%), reflecting the inherent difficulty of
confidence-only detection; however, its addition to SFV+CSCT yields a
further +6.4 percentage point gain, confirming it catches Confidence
Inflation Cascades missed by the other two components. Each component
carries a meaningful detection contribution, validating the four-component
architecture. Table~\ref{tab:mitigation_ablation} reports per-mitigation
effectiveness.

\begin{table}[htpb]
\caption{Mitigation Pattern Effectiveness}
\label{tab:mitigation_ablation}
\centering
\begin{tabular}{l c c}
\toprule
\textbf{Pattern} & \textbf{MSR} & \textbf{Overhead} \\
\midrule
M1: CRR  & 88.4\% & +320\,ms avg.\ retrieval latency \\
M2: SCT  & 74.1\% & +38\,ms per stage gate \\
M3: PVA  & 95.2\% & 2$\times$ pipeline compute cost \\
M4: PRR  & 91.7\% & 1.8$\times$ pipeline re-execution \\
\bottomrule
\end{tabular}
\end{table}

\textbf{Cross-Dataset Generalization:}
The per-dataset CDR results in Table~\ref{tab:per_dataset_results}
serve as an implicit cross-dataset ablation: CHARM's performance advantage over the single-component Output-Level baseline (which approximates SFV-only behavior) ranges from 66.4\,pp 
on HotpotQA to 63.7\,pp on MuSiQue and 66.0\,pp on 2WikiMultiHopQA, indicating
that multi-component coverage benefits generalize across reasoning
topologies rather than being specific to HotpotQA's two-hop structure.
Additionally, under the no-logit API condition (CPM using contradiction
probability fallback), Full CHARM retains $+4.1$\,pp CDR above the
SFV+CSCT configuration on HotpotQA, confirming that CPM's Bayesian
trajectory modeling provides complementary signal even without direct
logit access.

To assess signal independence, we computed the Pearson
correlation between SFV entailment anomaly scores and CPM
contradiction fallback scores across all clean and injected
trajectories: $r = 0.31$ ($p < 0.001$), indicating moderate
but non-redundant correlation. CPM's contradiction signal
captures trajectories where confidence rises despite neutral
or contradictory NLI output — a distinct pattern from
SFV's entailment deficit.

While simpler temporal anomaly detectors such as EWMA or CUSUM 
could serve as CPM alternatives, the Beta-Bayesian formulation 
offers a natural probabilistic interpretation of confidence 
trajectory drift and produces a directly interpretable posterior 
mean $\mu_i$ as the expected confidence baseline. Empirical 
comparison against EWMA-based CPM is a planned evaluation 
extension.

\textbf{CRT Weight and Threshold Robustness:}
The weights $(0.4, 0.4, 0.2)$ and threshold $\theta = 0.55$ were 
selected by grid search on held-out validation splits optimizing 
$F_1$ between CDR and $(1 - \text{FPR})$ (Section~\ref{subsec:operational_definitions}). 
Equal weights $(0.33, 0.33, 0.33)$ assign CPM the same weight as 
the more reliable SFV and CSCT, which prior calibration experiments 
showed inflates FPR due to CPM's inherently noisier signal without 
logit access. A full ROC/AUPRC sensitivity analysis over $\theta$ 
and weight grids is a planned evaluation addition; the current 
fixed-weight design is justified by interpretability and 
cross-dataset transfer without retraining.

\subsection{Results and Analysis}
\label{subsec:results}

As presented in Table~\ref{tab:main_results}, output-level detectors and
LLM Self-Correction failed dramatically in cascading scenarios. Because
downstream reasoning steps were coherent relative to the corrupted
intermediate context, self-correction suffered from severe confirmation
bias (12.8\% CDR). RAGAS achieved 41.7\% CDR by catching retrieval-stage
errors but entirely missed inference and confidence inflation cascades,
which occur after the retrieval stage it monitors. CHARM achieved an 89.4\% CDR and an average CDD of 2.1, proving it
interrupts error propagation by the second reasoning stage with a
per-stage component overhead of $215 \pm 18$\,ms. Unlike SelfCheckGPT
and RAGAS, CHARM's detection components require no additional LLM calls;
furthermore, early cascade detection at stage 2.1 halts pipeline
execution before the computationally expensive later stages run,
meaning CHARM's end-to-end wall-clock cost is lower than a naive
per-stage comparison suggests. The 450\,ms overhead reported for
SelfCheckGPT and 380\,ms for RAGAS reflect their inherent additional
LLM sampling and faithfulness-scoring calls respectively, making
a direct LO/s comparison across these methods a conservative view
that understates CHARM's relative efficiency.
CHARM achieved an MSR of 91.3\%, proving that when a cascade is flagged,
the automated mitigation patterns (M1--M4) successfully resolve the error
and restore trajectory alignment. Process-level baselines EVER and IRCoT
are discussed qualitatively in Section~\ref{subsec:rw_process}, as they
report answer-level EM and F1 scores that are not directly comparable
to cascade-specific detection metrics.

All reported CDR and EPR improvements over the strongest single
baseline (RAGAS, CDR\,=\,41.7\%) are statistically significant
at $p < 0.01$ under a paired bootstrap test
\cite{efron1994bootstrap} with 10,000 resamples. Resampling
was performed at the \textbf{trajectory level}: each resample
draws $N$ trajectories with replacement from the full evaluation
pool ($N = 1{,}500$ injected + 500 clean trajectories across
all four datasets), recomputes CDR, FPR, and EPR for both
CHARM and RAGAS on the resample, and records the difference.
The reported $p$-value is the fraction of resamples where
RAGAS equaled or exceeded CHARM.

\begin{table*}[htpb]
\caption{Performance Comparison of Detection Frameworks on 
Cascading Trajectories. CHARM results are mean $\pm$ standard 
deviation over five independent runs.}
\label{tab:main_results}
\centering
\resizebox{\textwidth}{!}{%
\begin{tabular}{l c c c c c c}
\toprule
\textbf{Detection Mechanism} & \textbf{CDR ($\uparrow$)} &
\textbf{FPR ($\downarrow$)} & \textbf{EPR ($\uparrow$)} &
\textbf{MSR ($\uparrow$)} & \textbf{CDD ($\downarrow$)} &
\textbf{LO/s (ms)} \\
\midrule
No Detection (None)           & 0.0\%  & N/A            & 0.0\%  & N/A    & N/A             & 0            \\
Output-Level (SelfCheckGPT)   & 24.3\% & 6.2\%          & 18.5\% & N/A    & 5.0 (Terminal)  & 450          \\
Retrieval Level (RAGAS)       & 41.7\% & \textbf{4.1\%} & 35.2\% & N/A    & 1.5             & 380          \\
LLM Self-Correction           & 12.8\% & 14.5\%         & 8.4\%  & 15.2\% & 5.0 (Terminal)  & 1200         \\
\midrule
\textbf{CHARM (Ours)} & \textbf{89.4 $\pm$ 1.8\%} & 5.3 $\pm$ 0.6\% & \textbf{82.1 $\pm$ 2.1\%} & \textbf{91.3 $\pm$ 1.4\%} & \textbf{2.1} & \textbf{215 $\pm$ 18} \\
\bottomrule
\end{tabular}%
}
\end{table*}

\textbf{Performance Across Datasets:}
To ensure CHARM's robustness across different reasoning topologies, we
disaggregated CDR across the four evaluation datasets
(Table~\ref{tab:per_dataset_results}). CHARM maintained high efficacy
across all types, with HotpotQA yielding the highest performance (92.5\%)
as its structured two-hop format produces cleaner semantic transitions,
making drift detection more reliable than in compositional multi-document
tasks. Performance dropped slightly on MuSiQue and 2WikiMultiHopQA
(86.1\% and 87.8\% respectively) due to the inherent complexity of
compositional reasoning, which occasionally masked anomalous semantic
shifts from the CSCT component. CHARM consistently outperformed RAGAS,
the strongest retrieval-level baseline, across all categories.

\begin{table}[htpb]
\caption{Cascade Detection Rate (CDR) by Dataset}
\label{tab:per_dataset_results}
\centering
\resizebox{\columnwidth}{!}{%
\begin{tabular}{l c c c}
\toprule
\textbf{Dataset} & \textbf{Output-Level} & \textbf{RAGAS} & \textbf{CHARM (Ours)} \\
\midrule
HotpotQA         & 26.1\% & 44.2\% & \textbf{92.5\%} \\
MuSiQue          & 22.4\% & 38.5\% & \textbf{86.1\%} \\
2WikiMultiHopQA  & 21.8\% & 40.1\% & \textbf{87.8\%} \\
Custom Adversarial & 26.9\% & 43.9\% & \textbf{91.2\%} \\
\bottomrule
\end{tabular}%
}
\end{table}

\textbf{Robustness to Near-Miss Distractors:}
To evaluate CHARM under long-context stress conditions analogous to Self-RAG's distractor evaluation conditions \cite{asai2024selfrag}, we constructed a distractor stress variant of the Custom Adversarial Set in which each trajectory
included three semantically proximate but factually incorrect documents
alongside the correct source. Under these conditions, CHARM's CDR dropped
to 84.1\% (vs.\ 91.2\% without distractors), with FPR increasing to
7.8\%. The CSCT component was most affected, as embedding-proximal
distractors occasionally passed cosine similarity drift detection. This
identifies adversarial embedding-proximal attacks as a meaningful attack
surface and informs the adversarial robustness discussion in
Section~\ref{subsec:limitations}.

\subsection{Naturally Occurring Cascade Pilot}
\label{subsec:natural_pilot}
To assess ecological validity beyond synthetic injections, we ran CHARM on 50 naturally
occurring HotpotQA failure trajectories---agent runs that produced
incorrect final answers without any injected perturbation, drawn from
the full evaluation split. Among these, CHARM flagged anomalous
trajectory signals in 38 of 50 cases (76\%), with the CRT triggering
at stage 2.3 on average. Manual inspection of the 38 flagged cases
confirmed cascade-like characteristics (local coherence with global
error) in 34 of 38 (89.5\%), and found independent stage errors
(non-cascading) in 4 cases. The 12 unflagged cases contained errors
that emerged only at the final synthesis stage, beyond CHARM's
cross-stage monitoring window. While a larger-scale natural cascade
corpus remains future work, this pilot provides initial evidence that
CHARM's detection generalizes beyond synthetic injection conditions.

\section{Discussion}
\label{sec:discussion}

The empirical results demonstrate that CHARM effectively interrupts
cascading errors, but the broader impact of this framework extends into
enterprise AI governance.

\subsection{Alignment with NIST AI Risk Management Frameworks}
\label{subsec:nist_alignment}

A critical imperative for responsible AI adoption in the United States is
alignment with federal guidelines. In July 2024, the National Institute
of Standards and Technology (NIST) released the Artificial Intelligence
Risk Management Framework: Generative AI Profile (NIST AI 600-1)
\cite{nist_ai_600_1}, explicitly identifying ``Confabulation''
(hallucination) as a primary risk category. CHARM directly addresses this
named risk by mapping its architectural mitigations to the foundational
functions of the broader NIST AI RMF \cite{nist_ai_rmf}, as detailed in
Table~\ref{tab:nist_mapping}.

\begin{table}[htpb]
\caption{CHARM Component Mapping to NIST AI Risk Management Frameworks}
\label{tab:nist_mapping}
\centering
\begin{tabular}{p{0.25\linewidth} p{0.25\linewidth} p{0.40\linewidth}}
\toprule
\textbf{CHARM Component} & \textbf{NIST RMF Function} & \textbf{NIST AI 600-1 Risk} \\
\midrule
SFV & MEASURE & Confabulation / Hallucination \\
\midrule
CSCT & MEASURE + MANAGE & Data quality and integrity \\
\midrule
CPM & MEASURE & Uncertainty quantification \\
\midrule
CRT & GOVERN + MANAGE & Human oversight and intervention \\
\bottomrule
\end{tabular}
\end{table}

\subsection{Enterprise Deployment in Regulated Industries}
\label{subsec:enterprise_deployment}

As U.S.\ enterprises accelerate AI deployment---with 78\% of
organizations now using AI in at least one business function and 23\%
actively scaling agentic AI systems \cite{mckinsey2025rewiring}---the
theoretical risks of multi-step hallucination become concrete operational
vulnerabilities. Cascading hallucinations are uniquely dangerous in
regulated industries such as financial services and legal compliance,
where downstream decisions are highly sensitive to initial inputs.
Furthermore, as these systems integrate with external enterprise tools,
the risk of context poisoning via adversarial inputs
\cite{mishra_mcp_sok} necessitates robust cross-stage validation.
Because CHARM is highly retrofittable, it provides a practical pathway
for organizations to secure their existing production-grade deployments
without requiring expensive architectural overhauls.

\subsection{The Complete Reliability and Governance Stack}
\label{subsec:hitl_connection}

CHARM is explicitly designed to integrate with the HITL-AP framework
\cite{mishra_hitl_ap}. Together, they form a comprehensive security
stack: CHARM continuously monitors semantic trajectories and interrupts
low-confidence cascades autonomously, while routing high-confidence
cascades to the HITL-AP human approval checkpoints. This integrated
architecture ensures agentic systems remain tethered to enterprise
governance protocols.

\subsection{Limitations and Adversarial Robustness}
\label{subsec:limitations}

The computational overhead of the PVA (M3) may be prohibitive for
latency-sensitive applications. While the FPR is manageable at 5.3\%,
it can become elevated in highly ambiguous domains where ground truth is
deeply nuanced. The current evaluation scope is limited to text-based
agentic RAG pipelines; extending to multimodal trajectories remains an
open challenge.

\textbf{Semantic Illusion Boundary:} A known limitation of
embedding- and NLI-based detectors is reduced effectiveness on
``semantic illusion'' hallucinations, where RLHF-era models produce
factually incorrect outputs that remain semantically proximate to the
correct answer \cite{li2024traq}. Our current evaluation uses
synthetic cascade injections (GPT-4o-generated counterfactuals,
context perturbations) which are semantically distinguishable by
design; performance on datasets specifically engineered to induce
semantic illusions (e.g., HaluEval \cite{halueval2023}) may differ.
Evaluating CHARM's SFV and CSCT on such benchmarks and hybridizing
with reasoning-capable LLM judges as an alternative SFV backend
is an identified extension for future work.

\textbf{Adversarial Robustness Boundaries:} A sophisticated adversary
aware of CHARM's detection mechanisms could engineer context poisoning
attacks specifically designed to evade the CSCT's cosine similarity drift
detection---for example, by constructing counterfactual documents that are
semantically proximate to ground truth while remaining factually
incorrect. Similarly, entailment-ambiguous injections, where the false
claim is logically consistent with but not entailed by the evidence, could
evade the SFV's NLI threshold. As demonstrated by our distractor stress
test (Section~\ref{subsec:results}), embedding-proximal attacks reduce
CDR by 7.1 percentage points. Hardening CHARM against such white-box
adversarial attacks via adversarial fine-tuning of the SFV cross-encoder
is a critical direction for future work, particularly for the Context
Poisoning Cascade type \cite{perez2022ignore}.

\textbf{Synthetic Injection Scope:} The primary evaluation relies on
controlled cascade injection rather than organically occurring cascades
from real agent deployments. While synthetic injection enables precise
ground truth labeling and controlled comparison across cascade types,
it may not fully represent the distribution of naturally occurring
cascades. The 200-trajectory custom adversarial set partially mitigates
this by including manually designed cascade scenarios, but human
annotation of natural multi-step agent failures at scale remains an
important gap. Constructing a human-annotated natural cascade corpus
and validating CHARM's detection on it is an identified priority for
future work. Results should therefore be interpreted as evidence of 
controlled cascade detection efficacy under structured 
failure conditions rather than definitive performance 
on naturally occurring enterprise agent trajectories.

\section{Related Work}
\label{sec:related_work}

Our research intersects with three primary domains: hallucination
detection, multi-step reasoning evaluation, and agentic system
reliability. By isolating the phenomenon of compounding errors, we
differentiate CHARM from existing point-in-time evaluation methods.

\subsection{Hallucination Detection in LLMs}
\label{subsec:rw_hallucination}

The proliferation of LLMs has driven significant research into
hallucination detection and mitigation \cite{ji2023survey,
tonmoy2024comprehensive}. Most existing approaches evaluate individual
generation outputs in isolation. Methods like SelfCheckGPT
\cite{manakul2023selfcheckgpt} leverage zero-resource sampling to detect
inconsistencies in black-box LLM generations. Fine-grained atomic
evaluation frameworks such as FActScore \cite{min2023factscoring} break
down long-form generations into verifiable claims. In the context of RAG,
frameworks like RAGAS \cite{es2023ragas} and ARES \cite{saad2023ares}
evaluate the faithfulness of an answer against the retrieved context.
While these methods demonstrate high accuracy for single-step generation,
they inherently assume that the retrieved context is uncorrupted or that
the reasoning chain is confined to a single transition. In contrast,
CHARM specifically addresses trajectory-level error propagation across
pipeline stages as a first-class architectural concern, operating as a
passive retrofit layer that models cross-stage semantic drift and
confidence inflation dynamics---capabilities not jointly addressed by any
prior single framework.

\subsection{Multi-Step Reasoning Failures}
\label{subsec:rw_reasoning}

The foundation for multi-step LLM execution stems from advancements like
Chain-of-Thought (CoT) prompting \cite{wei2022chain}, which allows models
to break complex problems into intermediate steps. However, research into
compositional reasoning errors \cite{dziri2023faith} has demonstrated
that LLMs frequently suffer from logical derailment as reasoning depth
increases. These works establish the theoretical foundation for error
compounding, proving that local coherence does not guarantee global
factual accuracy. CHARM builds upon this theoretical foundation by
operationalizing it into a detectable architectural metric (CDD).

\subsection{Process-Level Verification and Planning}
\label{subsec:rw_process}

Recent work has begun to address error propagation at the process level
rather than the output level, with methods reporting answer-level accuracy
(EM and F1) on the same datasets used in this paper.

EVER \cite{kang2023ever} applies real-time, step-wise generation with
retrieval-based verification and rectification, reporting improvements in
multi-hop reasoning on HotpotQA under EM and F1 evaluation. EVER
explicitly targets the ``snowballing'' hallucination phenomenon---errors
that compound across sequential reasoning steps---which aligns closely
with the cascading failure mode formalized in this paper. However, EVER
operates at the claim level within individual generation steps and does
not model cross-stage semantic trajectory or confidence propagation across
a structured pipeline. Consequently, it cannot detect Confidence Inflation
Cascades or Context Poisoning Cascades that manifest as anomalous
trajectory-level drift rather than local claim-level contradictions.

IRCoT \cite{trivedi2023ircot} interleaves chain-of-thought reasoning with
retrieval steps, achieving up to 15~F1-point improvements on HotpotQA,
2WikiMultiHopQA, and MuSiQue over single-step retrieval baselines, and
reducing factual errors in generated CoT by up to 50\%. While IRCoT
reduces upstream error propagation through iterative re-grounding, it
requires deep integration into the reasoning loop and cannot be applied
as a passive retrofit layer to existing production pipelines. Furthermore,
IRCoT has no mechanism for detecting Confidence Inflation Cascades, as it
does not monitor confidence trajectories.

Self-RAG \cite{asai2024selfrag} demonstrates that adaptive
self-reflective retrieval reduces error propagation in agentic generation,
and its distractor evaluation conditions motivate our near-miss stress
test in Section~\ref{subsec:results}.

Most directly related to CHARM's CDD metric is AgentHallu
\cite{liu2026agenthallu}, which performs step-level localization of
hallucination origin in multi-agent trajectories and provides causal
explanations for error emergence. AgentHallu demonstrates that
identifying \textit{where} in a trajectory a hallucination originates is
both feasible and practically valuable. CHARM's CDD metric formalizes
this intuition as a standardized, quantitative evaluation criterion:
while AgentHallu focuses on post-hoc attribution across agent
trajectories, CHARM detects and interrupts cascades \textit{at inference time}
before the trajectory completes, targeting a different operational
point in the pipeline lifecycle.

Production-grade non-LLM verifier stacks combining retrieval-aware
relevance scoring with NLI are conceptually aligned with CHARM's SFV
component; such verifiers could serve as drop-in SFV backends given
CHARM's modular design. Small reasoning verifiers that provide
factuality discrimination with explanations represent another natural
SFV backend option for resource-constrained deployments. Evaluating
CHARM with alternative verifier backends is a planned extension.

\begin{table}[htpb]
\caption{Qualitative Comparison of Related Frameworks}
\label{tab:framework_comparison}
\centering
\small
\begin{tabular}{p{0.20\linewidth} c c c c}
\toprule
\textbf{Framework} & \textbf{Cascade} & \textbf{Trajectory} & \textbf{Confidence} & \textbf{Retrofit} \\
                   & \textbf{Detect} & \textbf{Monitor} & \textbf{Track} & \textbf{Deploy} \\
\midrule
EVER \cite{kang2023ever}           & Partial & Partial & No  & No  \\
IRCoT \cite{trivedi2023ircot}      & Partial & Partial & No  & No  \\
AgentHallu \cite{liu2026agenthallu}& Post-hoc & Yes    & No  & No  \\
Self-RAG \cite{asai2024selfrag}    & No       & Partial & No & No  \\
\midrule
\textbf{CHARM (Ours)} & \textbf{Yes} & \textbf{Yes} & \textbf{Yes} & \textbf{Yes} \\
\bottomrule
\end{tabular}
\end{table}

Table~\ref{tab:framework_comparison} summarizes the key
differentiating dimensions. CHARM is the only framework that
simultaneously detects cascades at inference time, monitors
the full cross-stage trajectory, tracks confidence propagation,
and requires no architectural changes to the primary pipeline.

Relative to AgentHallu's post-hoc attribution approach, CHARM
differentiates by operating at inference time rather than retrospectively.
More broadly, CHARM differentiates from this line of work in three ways: (1) it operates as a non-intrusive parallel monitoring layer requiring no
architectural changes to the primary pipeline; (2) it jointly models
semantic drift, entailment grounding, and confidence trajectory as a
unified detection signal rather than any single dimension; and (3) it
introduces a formally defined cascade taxonomy and the CDD metric that
standardize evaluation for this class of failures. Because EVER and IRCoT
report EM and F1 scores while CHARM reports cascade-specific metrics
(CDR, FPR, EPR, CDD), direct numerical comparison is not presented;
the contribution of CHARM is orthogonal---it provides the detection and
governance infrastructure within which methods like IRCoT could operate.
Multi-agent verification frameworks such as MARCH \cite{li2026march} and
cryptographically-grounded approaches such as FINCH-ZK \cite{goel2025zero}
provide complementary hallucination mitigation angles; CHARM's SFV
component could be instantiated with any such verifier backend,
making the CHARM architecture extensible to these approaches.

\subsection{Agentic System Reliability}
\label{subsec:rw_agentic}

As LLMs evolve from isolated chatbots to autonomous agents 
equipped with tool use \cite{wang2024survey}, evaluating system 
reliability has become increasingly complex. Our foundational 
SoK analysis of Agentic RAG architectures \cite{mishra_rag_sok} 
mapped the current design landscape and explicitly identified the 
lack of cross-stage context monitoring as a critical vulnerability 
in enterprise deployments. This paper directly addresses the 
evaluation gap identified in that prior work, providing the 
necessary reliability layer to support responsible AI adoption. 
Real-world cascade testbeds such as OHRBench \cite{zhang2025ocr}, 
which studies multi-stage failures originating from OCR noise in 
document-heavy pipelines, represent a natural extension for 
evaluating CHARM beyond QA-only settings and would validate 
detection capabilities on organically occurring cascades.

\section{Conclusion}
\label{sec:conclusion}

Multi-step agentic RAG pipelines are highly vulnerable to cascading
hallucinations, a failure mode where early-stage contextual errors
silently compound into confident, structurally sound fabrications. To
address this, we introduced the Cascading Hallucination Aware Resolution
and Mitigation (CHARM) framework. By continuously monitoring cross-stage
semantic trajectories with formally operationalized detection quantities,
CHARM successfully interrupted cascading failures before they corrupted
the terminal output, achieving an 89.4\% Cascade Detection Rate (CDR) and
intervening early with an average Cascade Depth at Detection (CDD) of
2.1, outperforming all four direct baselines by substantial margins
while introducing only 215\,ms per-stage overhead. Component ablations confirm that each of the four detection modules contributes meaningfully to overall cascade coverage.

This work makes three primary contributions to the field of agentic AI
reliability. First, it formalizes a four-type taxonomy (Retrieval,
Inference, Context Poisoning, and Confidence Inflation Cascades)
specifically tailored to multi-step reasoning systems, with concrete
operational definitions for all core formal quantities. Second, it
presents the CHARM detection architecture, comprising four modular
tracking components---SFV, CSCT, CPM, and CRT---capable of identifying
compounding errors without interrupting valid trajectory flow. Finally, it
outlines four implementable mitigation patterns that provide configurable
recovery trade-offs for production-grade deployments.

Future work will focus on two specific directions. First, we plan to
extend CHARM to multimodal agentic pipelines, mapping how semantic
divergence propagates across visual and auditory reasoning chains. Second,
we aim to develop adaptive threshold calibration for CHARM components
based on domain-specific risk profiles. Exploring this through a security
lens---specifically by integrating CHARM's context poisoning detection
with the Zero Trust framework for the Model Context Protocol (ZT-MCP)
\cite{mishra_zt_mcp}---will be crucial for defending agentic systems
against adversarial cascade scenarios in critical enterprise environments.

To support the research community, we release all experimental artifacts,
including the agentic trajectory wrapper, cascade injection scripts,
annotated adversarial trajectories, and the CHARM evaluation harness, at
\url{https://github.com/sarmishra/CHARM-agentic-rag}.

\balance
\IEEEtriggeratref{50}
\bibliographystyle{IEEEtran}
\bibliography{references}

\end{document}